\documentclass[10pt,twocolumn,letterpaper]{article}

\usepackage[pagenumbers]{cvpr} 

\usepackage{graphicx}
\usepackage{amsmath}
\usepackage{amssymb}
\usepackage{booktabs}
\usepackage[table,dvipsnames]{xcolor}
\usepackage{bm}
\usepackage{array}
\usepackage{multirow}
\usepackage{times}
\usepackage{epsfig}
\usepackage{ulem}
\usepackage[accsupp]{axessibility}
\newcommand{\xs}[1]{\bm{x}_s^\text{#1}}
\newcommand{\ys}[1]{\bm{y}_s^\text{#1}}

\newcommand{\algoname}{MM2D3D}

\usepackage[pagebackref,breaklinks,colorlinks]{hyperref}

\usepackage[capitalize]{cleveref}
\crefname{section}{Sec.}{Secs.}
\Crefname{section}{Section}{Sections}
\Crefname{table}{Table}{Tables}
\crefname{table}{Tab.}{Tabs.}

\def\cvprPaperID{4} 
\def\confName{CVPR}
\def\confYear{2023}

\begin{document}

\title{Exploiting the Complementarity of 2D and 3D Networks to Address Domain-Shift in 3D Semantic Segmentation}

\author{Adriano Cardace\qquad Pierluigi Zama Ramirez\qquad Samuele Salti\qquad Luigi Di Stefano\\
Department of Computer Science and Engineering (DISI)\\
University of Bologna, Italy\\
{\tt\small \{adriano.cardace2\}@unibo.it}
}
\twocolumn[{
\renewcommand\twocolumn[1][]{#1}
\maketitle
\begin{center}
    \includegraphics[width=0.8\textwidth]{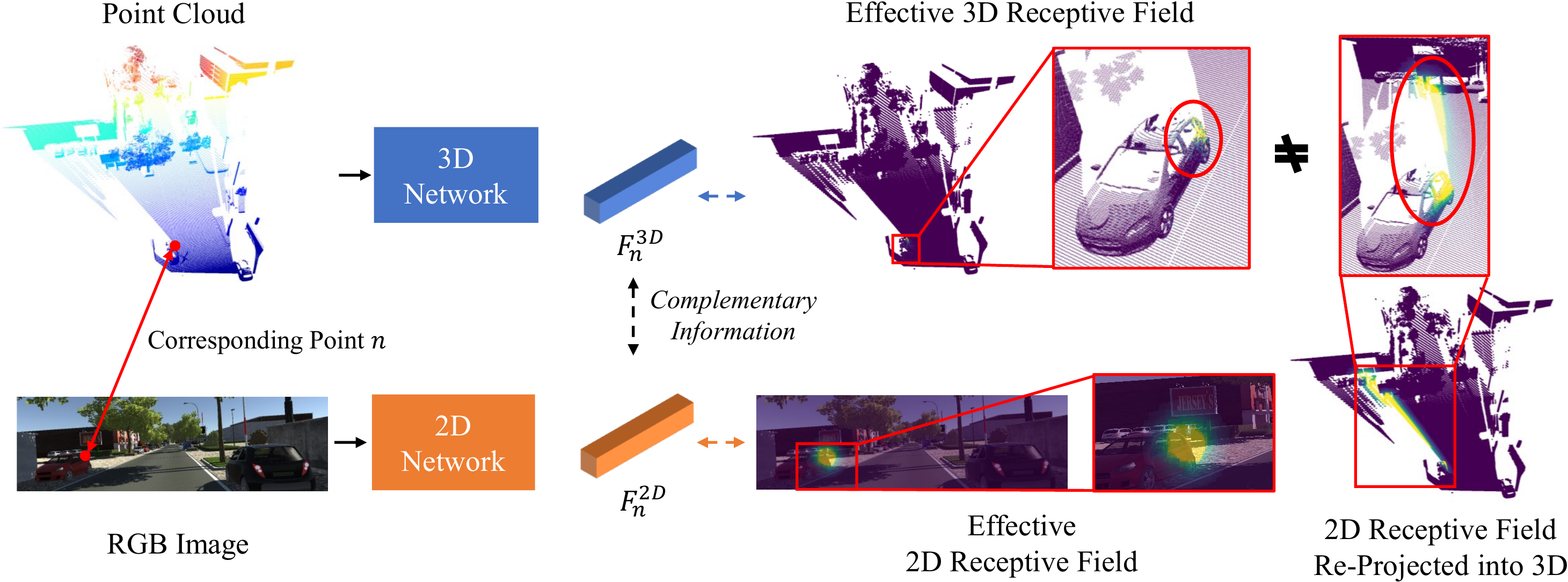}
    \label{fig:teaser}
\end{center}
\vspace{-0.2cm}
\small \hypertarget{fig:teaser}{Figure 1.} 3D (top) and 2D (bottom) networks processing point clouds and images of the same scene extract features that contain complementary information. Indeed, 2D and 3D effective receptive fields \cite{luo2016understanding} centered on a point $n$ focus on different portions of the scene, i.e., 2D or 3D neighborhoods respectively. Thus, corresponding features have different content by construction. We exploit this property to reduce the domain gap in 3D semantic segmentation.  
\vspace{0.38cm}
}]

\setcounter{figure}{1}

\maketitle

\begin{abstract}
3D semantic segmentation is a critical task in many real-world applications, such as autonomous driving, robotics, and mixed reality.
However, the task is extremely challenging due to ambiguities coming from the unstructured, sparse, and uncolored nature of the 3D point clouds. A possible solution is to combine the 3D information with others coming from sensors featuring a different modality, such as RGB cameras.
Recent multi-modal 3D semantic segmentation networks exploit these modalities relying on two branches that process the 2D and 3D information independently, striving to maintain the strength of each modality. In this work, we first explain why this design choice is effective and then show how it can be improved to make the multi-modal semantic segmentation  more robust to domain shift. Our surprisingly simple contribution achieves state-of-the-art performances on four popular multi-modal unsupervised domain adaptation  benchmarks, as well as  better results in a domain generalization scenario.
\end{abstract}

\section{Introduction}
\label{sec:intro}
3D semantic segmentation is a critical task in many real-world applications, such as autonomous driving and robotics. It involves assigning labels to 3D points in a point cloud based on their semantic meaning.
However, this task can be extremely challenging due to ambiguities coming from the unstructured, sparse, and uncolored nature of the 3D point clouds.
Fortunately, combining 3D information with others coming from sensors with a different modality, such as RGB cameras, can help to address these shortcomings. Indeed, by combining multi-modal data, we can leverage the strengths of each modality to produce more comprehensive and accurate segmentations.
For example, in  autonomous driving scenarios, RGB cameras and LiDARs are commonly used together. RGB cameras provide dense, colored, and structured information, but they may fail in dark lighting conditions. On the other hand, LiDARs are robust to light conditions, but the point clouds present the problems highlighted above.
By combining these two modalities, we can obtain a richer understanding of the environment and make more robust and precise 3D segmentations.

Several recent approaches for multi-modal 3D semantic segmentation \cite{jaritz2020xmuda, jaritz2022cross, dscml, yan20222dpass, shin2022mm} leverage a peculiar two-branch 2D-3D architecture, in which images are processed by a 2D convolutional network, e.g., ResNet \cite{resnet}, while point clouds by a 3D convolutional backbone, e.g., SparseConvNet \cite{sparseconv}.
By processing each modality independently, each of the two branches focuses on extracting features from its specific signal (RGB colors or 3D structure information) that can be fused effectively due to their inherent complementarity in order to produce a better segmentation score.
Indeed, averaging logits from the two branches  provides often an improvement in performance, e.g., a mIoU gain from 2\% to 4\% in almost all experiments in \cite{jaritz2020xmuda}.
Although we agree that each modality embodies specific information, such as color for images and 3D coordinates for point clouds, we argue that the complementarity of the features extracted by the two branches is also tightly correlated to the different information processing machinery,  i.e., 2D and 3D convolutions, which makes networks focusing on different areas of the scene with different receptive fields.
Indeed, in Fig. \hyperref[fig:teaser]{1}, given a point belonging to the red car,  we visualize the effective receptive field \cite{luo2016understanding} of the 2D and 3D networks (red ellipses). As we can clearly see from the receptive fields in the right part of the figure, the features extracted by the 3D network mainly leverage points in a 3D neighborhood, i.e., include points of the car surface. In contrast, the features extracted by the 2D network look at a neighborhood in the 2D projected space, and thus they depend  also on pixels of the building behind the car, which are close in image space but far in 3D.
We argue that this is one of the main reasons why the features from the two branches can be fused so effectively.
Based on the above intuition, we propose to feed 3D and RGB signals to \emph{both} networks as this should not hinder the complementarity of their predictions, with the goal of making the network more robust to the change of distributions between the training and the test scenarios. This problem is typically referred to \textit{Domain shift} in the literature.
Feeding both branches with both modalities would make: i) the 2D network more robust to domain shifts, as depth information (z coordinates of point clouds projected into image space) is more similar across different domains, as shown in several papers \cite{dada, saha2021learning, geoguided, multichannel, atdt, cardace2022plugging}; ii) the 3D network more capable of adapting to new domains thanks to RGB information associated with each point which allows learning better semantic features for the target domain, when this is available, using Unsupervised Domain Adaptation (UDA) approaches.
%
Thus, we propose a simple architecture for multi-modal 3D semantic segmentation consisting of a 2D-3D architecture with each branch fed with both RGB and 3D information. Despite its simplicity, our proposal achieves state-of-the-art results in multi-modal UDA benchmarks, surpassing competitors by large margins, as well as significantly better domain generalization compared to a standard 2D-3D architecture \cite{jaritz2022cross}.
Code available at {\small{\url{ https://github.com/CVLAB-Unibo/MM2D3D}}.}
Our contributions are:
\begin{itemize}
    \item shining a light on the intrinsic complementarity of recent multi-modal 3D semantic segmentation networks  based on 2D-3D branches;
    \item proposing a simple yet remarkably effective baseline that injects depth cues into the 2D branch and RGB colors into the 3D branch while preserving the complementarity of predictions;
    \item our network achieves state-of-the-art results in popular UDA benchmarks for multi-modal 3D semantic segmentation and surpasses standard 2D-3D architectures in domain generalization.
\end{itemize}

\section{Related works}
\label{sec:related}

\textbf{Point Cloud Semantic Segmentation.}
3D data can be represented in several ways such as point clouds, voxels, and meshes, each with its pros and cons. 
Similarly to pixels in 2D, voxels represent 3D data as a discrete grid of the 3D space. This representation allows using convolutions as done for images. However, performing a convolution over the whole 3D space is memory intense, and it does not consider that many voxels are usually empty. Some 3D CNNs \cite{riegler2017octnet, tatarchenko2017octree} rely on OctTree \cite{meagher1982geometric} to reduce the memory footprint but without addressing the problem of manifold dilation. SparseConvNet \cite{sparseconv} and similar implementations \cite{choy20194d} address this problem by using hash tables to convolve only on active voxels, allowing the processing of high-resolution point clouds with only one point per voxel. Aside from cubic discretization, some approaches \cite{zhu2021cylindrical, zhang2020polarnet} employ cylindrical voxels. Other methods address the problem with sparse point-voxel convolutions \cite{tang2020searching}.
Differently, point-based networks process directly each point of a point cloud.
PointNet++ \cite{qi2017pointnet++} extract features from each point, and then extract global and local features by means of max-pooling in a hierarchical way. Many improvements have been proposed in this direction, such as continuous convolutions \cite{thomas2019kpconv}, deformable kernels \cite{thomas2019kpconv} or lightweight alternatives \cite{hu2020randla}.
In this work, we select SparseConvNet \cite{sparseconv} as our 3D network as done by other works in the field \cite{jaritz2020xmuda, dscml, yan20222dpass, shin2022mm} since it is suitable for 3D semantic segmentation of large scenes.

\begin{figure*}[t]
    \centering
    \scalebox{0.7}{
    \includegraphics[clip, trim=1cm 6cm 1cm 4cm, width=\linewidth]{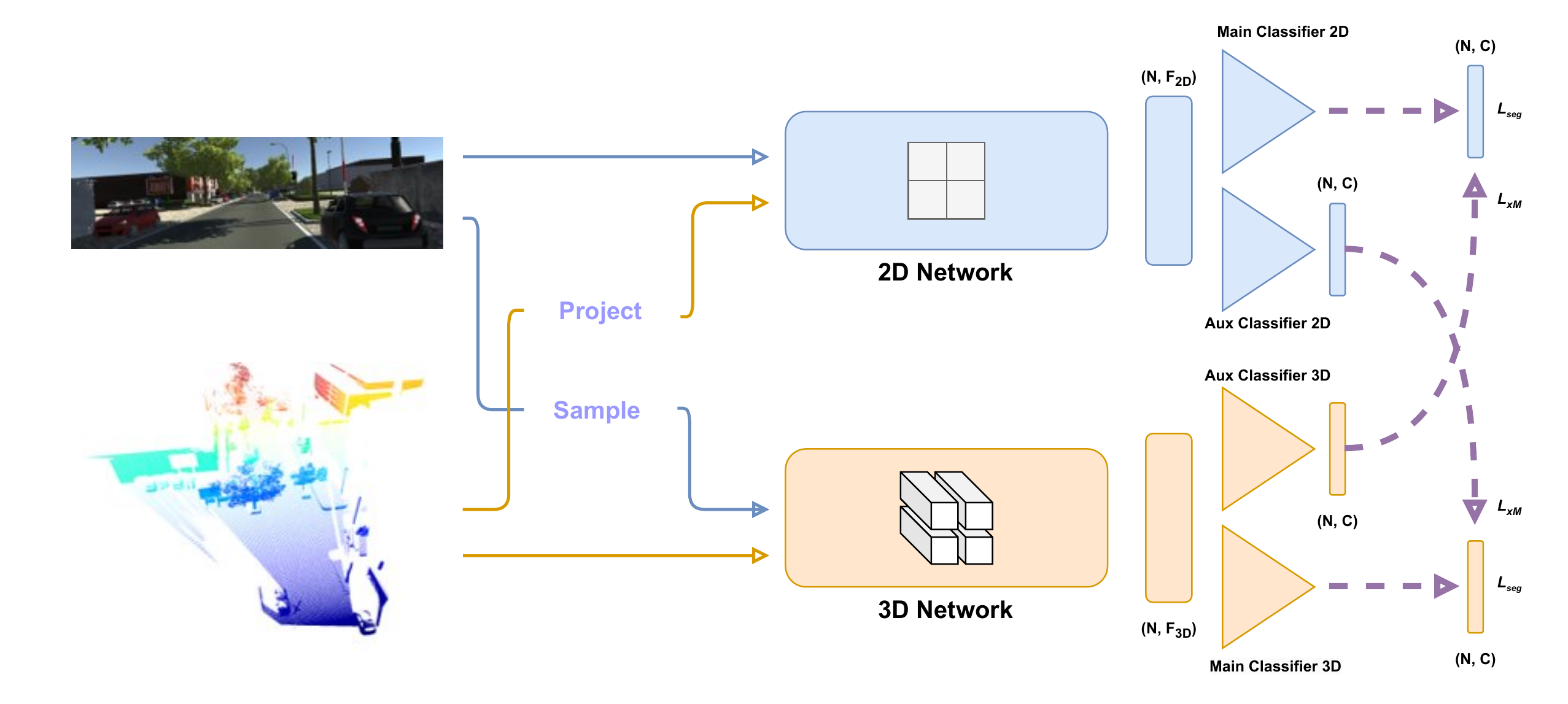}}
    \caption{\textbf{Framework overview.} The RGB image and the sparse depth map obtained from the projection of the corresponding point cloud are fed to a custom 2D architecture to extract point-wise features. The same point cloud and sampled colors from the RGB image are given in input to the 3D Network. Then, two main classifiers output the main predictions to be used at test time. Moreover, two auxiliary classifiers are used at training time only to allow the exchange of information across branches.}
    \label{fig:framework}
\end{figure*}

\textbf{Multi-Modal Learning.}
Exploiting multiple modalities to learn more robust and performant networks is a well-studied field in the literature \cite{ngiam2011multimodal, mm_survey}. Among them, several approaches address the problem of semantic segmentation exploiting RGB and 3D structure information, either with the final goal of segmenting images, e.g., RGB-D networks \cite{hazirbas2017fusenet,valada2020self} or point clouds, e.g., LiDAR + RGB approaches \cite{guo2020deep,krispel2020fuseseg,yan20222dpass}.
To speed-up research in this promising field, several datasets have been collected \cite{semanticKITTI,a2d2,nuscenes,gaidon2016virtual} with 3D point clouds, images, and annotations for tasks such as 3D object detection or 3D semantic segmentation.
Recently, some multi-modal methods \cite{jaritz2020xmuda, dscml, yan20222dpass, shin2022mm} show that a framework composed of a 2D and a 3D network can obtain very good performance in popular 3D segmentation benchmarks when averaging the scores coming from the two branches. This result is ascribed to the complementarity of the predictions due to the different modalities processed by each branch (either RGB or point clouds). In this paper, we analyze the improvement obtained by fusing the scores, and we argue that it mainly depends on the fact that the two networks extract complementary features because of the different receptive fields of the 2D and 3D networks. Based on this intuition, we propose a simple yet effective modification of the 2D-3D framework, that consists of providing both modalities as input to both branches.

\textbf{Unsupervised Domain Adaptation.}
Unsupervised Domain Adaptation is the research field that investigates how to transfer knowledge learned from a source annotated domain to a target unlabelled domain \cite{wang2018deep}.
In the last few years, several UDA approaches have been proposed for 2D semantic segmentation, using strategies such as style-transfer \cite{cycada,ipas,dcan,Chang_2019, Choi_2019, Murez_2018, Zhang_2018, pizzati, bdl,ltir}, adversarial training to learn domain-invariant representations \cite{hoffman2016fcns, adaptsegnet, patch, fada, joint_adversarial, adversarial_perturbation, michieli, ADVENT, stuffAndThings} or self-training \cite{cbst, zhang2021prototypical, cardace2022shallow, hoyer2022hrda, hoyer2022daformer}.
Recently, some works demonstrated the effectiveness of using depth information to boost UDA for 2D semantic segmentation \cite{dada, saha2021learning, geoguided, multichannel, atdt, cardace2022plugging, atdt2}. 
In our work, we take inspiration from these findings, and we feed the projected point cloud in input to also to the 2D network, considering depth as a rich source of information robust to the domain shift.
Recently some works address UDA also for semantic segmentation of point clouds \cite{rist2019cross, jiang2021LiDARnet, wu2019squeezesegv2, saleh2019domain, alonso2020domain, langer2020domain, zhao2021epointda, yi2021complete, piewak2019analyzing}. 
Very recently, some works have addressed the challenging multi-modal 3D semantic segmentation task \cite{jaritz2020xmuda, jaritz2022cross, dscml, shin2022mm}.
XMUDA \cite{jaritz2020xmuda} is the first work that focuses on UDA in the above setting, it defines a new benchmark and a baseline approach to adapt to a new target domain with an unsupervised cross-modal loss. \cite{jaritz2022cross} extend it, by proposing a more solid and comprehensive benchmark. DsCML \cite{dscml} also extends XMUDA deploying adversarial training to align features across modalities and domains.
In our work, we address the same multi-modal UDA scenarios introduced in \cite{jaritz2022cross}, and we propose a simple yet effective architecture that is more robust to domain shift and can be adapted to new unlabelled target domains.
Our framework, depicted in \cref{fig:framework}.

\section{Method}
\textbf{Setup and Notation.}
We define input source samples $\{\bm{x}_s^{2D}, \bm{x}_s^{3D}\} \in \mathcal{S}$ and target samples $\{\bm{x}_t^{2D}, \bm{x}_t^{3D}\} \in \mathcal{T}$, with $\bm{x}^{2D}$ being the 2D RGB image and $\bm{x}^{3D}$ the corresponding point cloud, with 3D points in the camera reference frame.
Note that $\bm{x}^{3D}$ contains only points visible from the RGB camera, assuming that the calibration of the two sensors is available for both domains and does not change over time.
We assume the availability of annotations $\bm{y}_s^{3D}$ only for the source domain for each 3D point. 
When tackling the UDA scenario, we also have at our disposal the unlabeled samples from the target domain.
Our goal is to obtain a point-wise prediction $N \times C$ for $\bm{x}_t^{3D}$, with $N$ and $C$ being the number of points of the target point cloud and the number of classes, respectively.

\begin{figure}[t]
    \centering
    \includegraphics[width=0.7\linewidth]{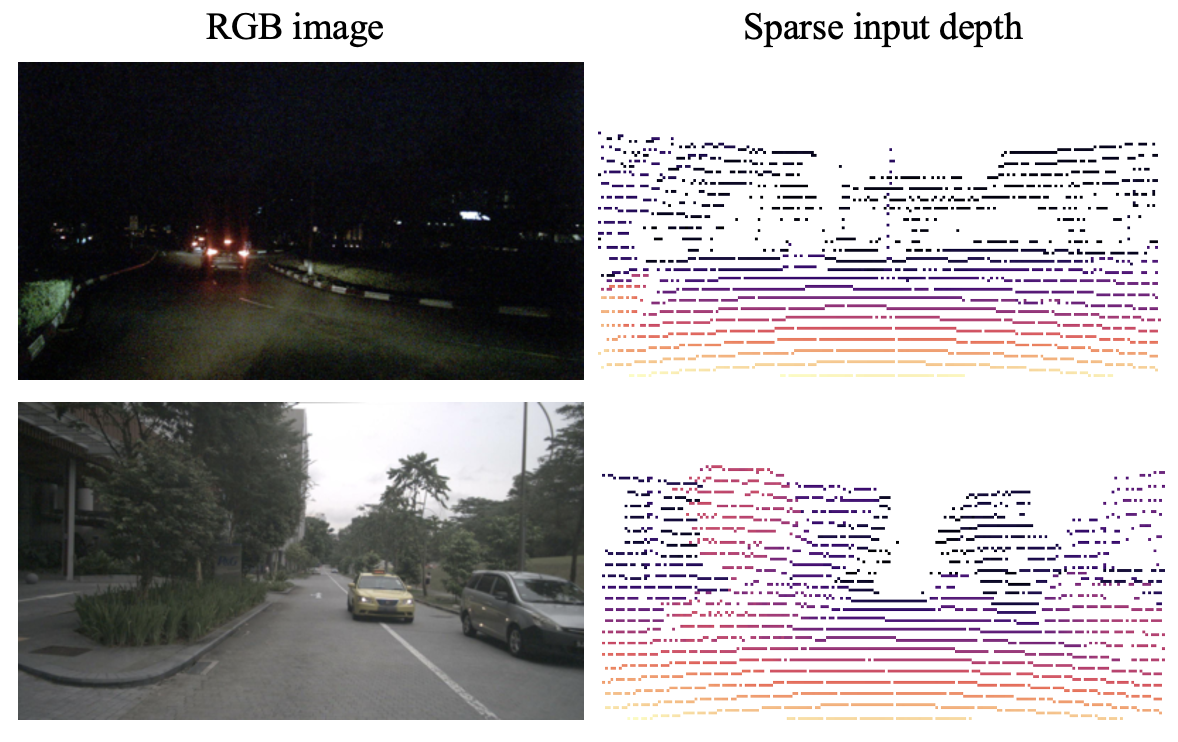}
    \caption{Depth comparison during daylight or night. Differently, from the RGB image (left column), a sparse depth map obtained by projecting a LiDAR scan into the image plane is not affected by the light conditions.\vspace{-0.5cm}}
    \label{fig:day_vs_night}
\end{figure}

\subsection{Base 2D/3D Architecture}
\label{subsec:baseline}
We build our contributions upon the two independent branches (2D and 3D) architecture proposed in \cite{jaritz2022cross}. 
The 2D branch processes images to obtain a pixel-wise prediction given $\bm{x}^{2D}$ and it consists of a standard 2D U-Net\cite{unet}.
On the other hand, the 3D branch takes in input point clouds to estimate the class of each point of $\bm{x}^{3D}$ and it is implemented as a 3D sparse convolutional network \cite{sparseconv}.
Thanks to the fact that 2D-3D correspondences are known, 3D points can be projected into the image plane to supervise the 2D branch, as supervision is provided only for the sparse 3D points. 
We denote the 3D semantic labels projected into 2D with the symbol $\bm{y}^{3D\rightarrow 2D}$.
As argued by \cite{jaritz2022cross}, such design choice allows one to take advantage of the strengths of each input modality, and final predictions can be obtained by averaging the outputs of the two branches to achieve an effective ensemble.
In our work, we adopt the same framework, and we give an intuitive explanation of why this design choice is particularly effective. In particular, we reckon that the two predictions are complementary not only for the input signals being different but also for the fact the two branches focus on different things to determine their final predictions. Indeed, 3D convolutions produce features by looking at points that are close in the 3D space, while the 2D counterparts focus on neighboring pixels in the 2D image plane. 
Therefore, given corresponding 2D and 3D points, the two mechanisms implicitly produce features containing complementary information.
In the right part of Fig. \hyperref[fig:teaser]{1} we visualize the Effective Receptive Fields (ERF) \cite{luo2016understanding} of a 2D U-Net with backbone ResNet34\cite{resnet} and of a 3D U-Net with backbone SparseConvNet\cite{sparseconv}.
It is worth  highlighting that we do not focus on the theoretical yet on the effective receptive field, which is computed by analyzing the real contribution of each input point to the final prediction (the hotter the color intensity in the visualization, the larger the point contribution).
Comparing the re-projected 2D ERF into 3D and the 3D ERF we can clearly appreciate that the 2D network focuses on sparse 3D regions, i.e., from the car to the building in the background, while the 3D counterpart reasons on a local 3D neighborhood (only car points). 
With this intuition in mind, we argue that by feeding the RGB signal to the 3D network, and the 3D information to the 2D backbone, we would still obtain complementary features that can be effectively fused together.
Moreover, it is well-known that employing depth information as input to 2D segmentation networks can make it more robust to domain shift \cite{hazirbas2017fusenet, cardace2022plugging}. At the same time, we posit that the 3D network with RGB information may be able to extract better semantic features.
Differently from previous approaches that employ two independent architectures, based on the above considerations, we propose our multi-modal, two-branch framework named \algoname{}.
In \cref{subsec:depth_encoder} we show how a point cloud can be used to obtain a stronger and more suitable input signal for the 2D network. Similarly, in \cref{subsec:rgb_encoder} we describe our multi-modal 3D network.

\begin{figure}[t]
    \centering
    \includegraphics[width=0.65\linewidth]{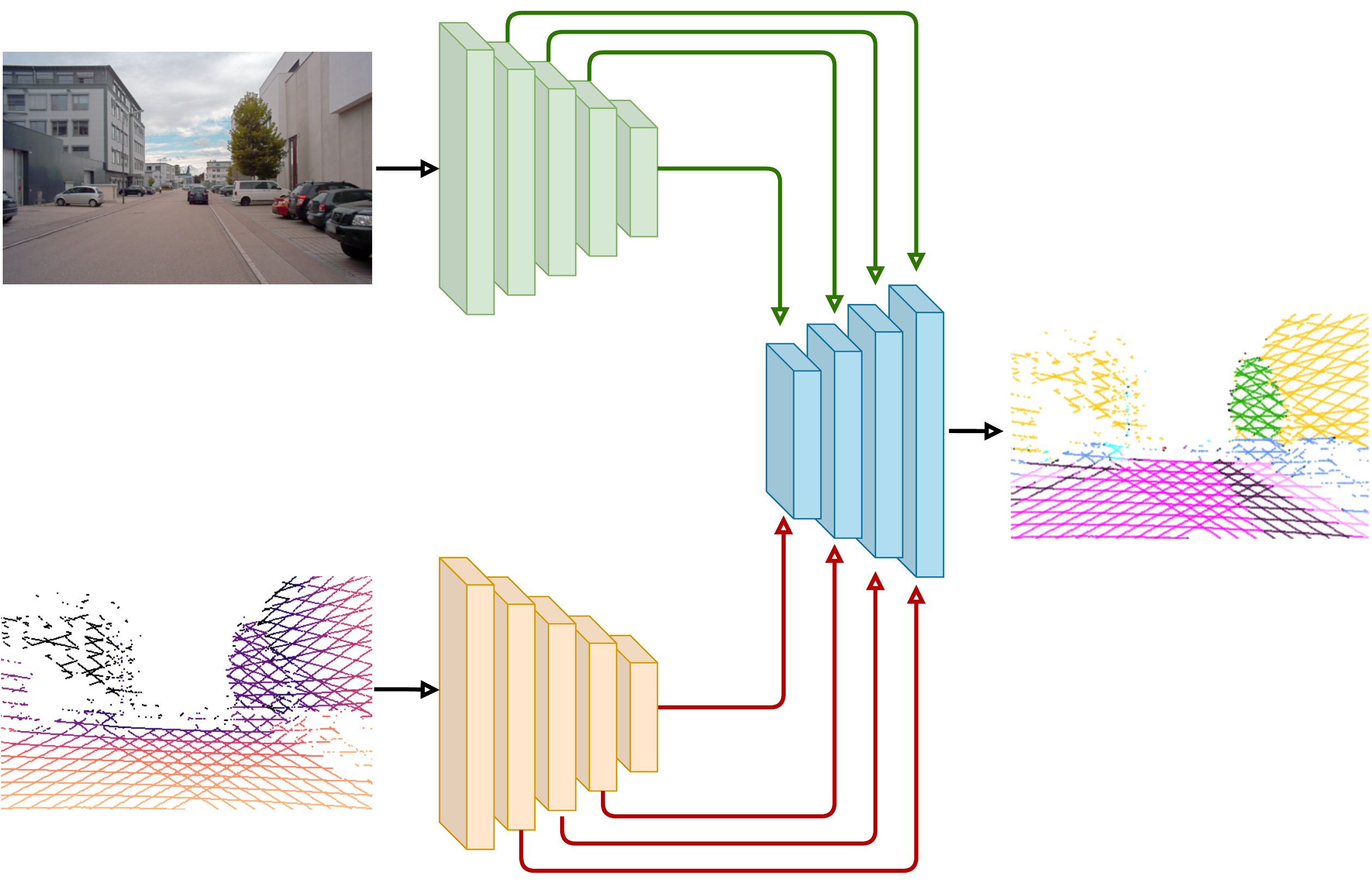}
    \caption{2D Network of our framework. It is composed of a depth encoder and an RGB encoder to process the two inputs independently. The segmentation decoder leverages the multi-scale features of both encoders to predict semantic segmentation labels.}
    \label{fig:net_2d}
\end{figure}

\subsection{Depth-based 2D Encoder}
\label{subsec:depth_encoder}
In this section, we focus on how we can use point clouds to make a  2D segmentation architecture more robust to domain shift. 
Inspired by \cite{hazirbas2017fusenet, cardace2022plugging}, we propose to use depth maps as an input signal that is less influenced by the domain gap.
As we can observe from the two depth maps in \cref{fig:day_vs_night}, it is hard to understand which one was captured during day or night. 
At the same time, some objects such as the car can be distinguished by only looking at depths (bottom right of the second depth map). Thus, depth maps provide useful hints to solve the task of semantic segmentation. 
Given these considerations, we argue that exploiting such invariant information may alleviate the domain shifts and can be used to extract discriminative features for the segmentation task.
At a first glance, injecting 3D cues into the 2D branch may seem redundant as the 3D network already has the capability to reason on the full 3D scene. 
However, given that the two networks have very different receptive fields, we can exploit such additional and useful information without the risk of hindering the complementarity of the two signal streams.
Assuming point clouds expressed in the camera reference frame and the availability of the intrinsic camera matrix, we can project the original 3D point cloud to obtain a sparse depth map.
In practice, the value of the $z$ axis is assigned to the pixel coordinate $(u,v)$ obtained by projecting a 3D point into the image plane.
Similarly to \cite{hazirbas2017fusenet}, to process both inputs, we modify the 2D encoder of the 2D U-Net architecture by including an additional encoder to process the sparse depth maps obtained from the point cloud. As can be seen in \cref{fig:net_2d}, the two streams i.e. one for the RGB image and the other for the sparse depth map, are processed independently.
Then, the concatenated depth and RGB features are processed by a decoder, composed of a series of transposed convolutions and convolutions in order to obtain semantic predictions of the same size as the input image. Moreover, features from layers of $\frac{1}{2}$ to $\frac{1}{16}$ of the input resolution are concatenated using skip connections with the corresponding layer of the decoder.
This simple design choice allows semantic predictions to be conditioned also on the input depth signal, without altering the RGB encoder that provides useful classification features. Furthermore, without altering the RGB encoder, we can take advantage of a pre-trained architecture on ImageNet \cite{imagenet} as done by our competitors.

\subsection{RGB Based 3D Network}
\label{subsec:rgb_encoder}
In this work, we focus on the 3D convolutional network, SparseConvNet \cite{sparseconv}, as it can segment large scenes efficiently. 
In this network, the initial point cloud is first voxelized such that each 3D point is associated with only one voxel. Then, rather than processing the entire voxel grid, these models work with a sparse tensor representation ignoring empty voxels for the sake of efficiency. The network associates a feature vector to each voxel, and convolutions calculate their results based on these features. A standard choice for the voxel features is to simply assign to it a constant value, i.e., 1.
Although these strategies have been shown to be effective \cite{yan20222dpass, saltori2022cosmix}, the feature vector can be enriched to make it even more suitable for semantic segmentation.
Based on our intuition of the different receptive fields, we can borrow information from the other modality to improve the performance of each branch, still preserving 2D-3D feature complementarity.
Thus, we use RGB colors directly as features for each voxel of the SparseConvNet. 
Moreover, we design a simple yet effective strategy to let the 3D network decide whether to use or not this information.
More specifically, the original RGB pixel values are fed to a linear layer that predicts a scalar value $\alpha$ to be multiplied by the color vector. 
For instance, learning this scaling could be useful in the UDA scenario, where we can train on unlabelled target samples, to discard RGB colors in case they do not provide any useful information, e.g., dark pixels in images acquired at night time.

\subsection{Learning Scheme}
\label{subsec:crossmodal}

\textbf{Supervised Learning.}
Given the softmax predictions of the 2D and 3D networks, $P_\text{2D}$ and $P_\text{3D}$, we supervise both branches using the cross-entropy loss on the source domain:
\begin{align}\label{eq:SegLoss}
\mathcal{L}_{\text{seg}}(\bm{x}_s, \ys{}) = - \frac{1}{N}\sum_{n=1}^N \sum_{c=1}^C \bm{y}_s^{(n, c)} \log \bm{P}_{\bm{x}_s}^{(n, c)}
\end{align}
with $(\bm{x}_s, \bm{y}_s)$ being either $(\xs{2D}, \bm{y_s}^{3D\rightarrow 2D})$ or $(\xs{3D}, \bm{y_s}^{3D})$.

\textbf{Cross-Branch Learning.}
To allow an exchange of information between the two branches, \cite{jaritz2022cross} and \cite{dscml} add an auxiliary classification head to each one. The objective of these additional classifiers is to mimic the other branch output. 
The two auxiliary heads estimate the other modality output: 2D mimics 3D ($P_{\text{2D} \to \text{3D}}$) and 3D mimics 2D ($P_{\text{3D} \to \text{2D}}$). 
In practice, this is achieved with the following objective:
\begin{align}\label{eq:CMLLoss}
\mathcal{L}_{\text{xM}}(\bm{x}) &= \bm{D}_\text{KL}(\bm{P}_{\bm{x}}^{(n, c)} || \bm{Q}_{\bm{x}}^{(n, c)}) \\
&= - \frac{1}{N} \sum_{n=1}^N \sum_{c=1}^C \bm{P}_{\bm{x}}^{(n, c)} \log \frac{\bm{P}_{\bm{x}}^{(n, c)}}{\bm{Q}_{\bm{x}}^{(n, c)}} \notag
\end{align}
with $(\bm{P}, \bm{Q}) \in  \{(\bm{P}_\text{2D}, P_{\text{3D} \to \text{2D}}), (\bm{P}_\text{3D}, P_{\text{2D} \to \text{3D}})\}$ where $\bm{P}$ is the distribution from the main classification head which has to be estimated by $\bm{Q}$.
Note that in \cref{eq:CMLLoss}, $x$ can belong to either $\mathcal{T}$ or $\mathcal{S}$. This means that, in the UDA scenario, \cref{eq:CMLLoss} can also be optimized for $\mathcal{T}$, forcing the two networks to have consistent behavior across the two modalities for the target domain as well without any labels.

\textbf{Self-Training.}
Only in the UDA scenario, where unlabelled target samples are available, as done by \cite{jaritz2022cross}, we perform one round of Self-Training \cite{cbst} using pseudo-labels \cite{pseudo_labels}. 
Specifically, after training the model with \cref{eq:SegLoss} for the source domain and \cref{eq:CMLLoss} on both domains, we generate predictions on the unlabeled target domain dataset to be used as pseudo ground truths, $ \bm{\hat{y}}_t$.
Following \cite{jaritz2022cross}, we filter out noisy pseudo-labels by considering only the most confident predictions for each class. Then, we retrain the framework from scratch the model minimizing the following objective function:

\begin{align}
\label{eq:total_loss}
\mathcal{L} &= \mathcal{L}_{\text{seg}}(\bm{x}_s, \ys{}) + \lambda_t \mathcal{L}_{\text{seg}}(\bm{x}_t, \bm{\hat{y}}_t)\\
&+ \lambda_{xs} \mathcal{L}_{\text{xM}}(\bm{x_s}) + \lambda_{xt} \mathcal{L}_{\text{xM}}(\bm{x_t})   \notag
\end{align}

\begin{table*}[t]
    \centering{}
    \scalebox{0.65}{
    \begin{tabular}
    {cl>
    {\centering}p{1cm}>
    {\centering}p{1cm}>
    {\centering}p{1cm}>
    {\centering}p{0.1cm}>
    {\centering}p{1cm}>
    {\centering}p{1cm}>
    {\centering}p{1cm}>
    {\centering}p{0.1cm}>
    {\centering}p{1cm}>
    {\centering}p{1cm}>
    {\centering}p{1cm}>
    {\centering}p{0.1cm}>
    {\centering}p{1cm}>
    {\centering}p{1cm}>
    {\centering}p{1cm}>
    {\centering}p{0.1cm}>
    {\centering}p{1cm}>
    {\centering}p{1cm}>
    {\centering}p{1cm}>
    {\centering}p{0.1cm}>
    {\centering}p{1cm}>
    {\centering}p{1cm}>
    {\centering}p{1cm}
    }
    \hline 

    \multirow{2}{*}{Modality} & \multirow{2}{*}{~~~~~~~~~~~~Method} & \multicolumn{3}{c}{\cellcolor{blue!25}USA \textrightarrow{} Singapore} &  & \multicolumn{3}{c}{\cellcolor{YellowOrange}Day \textrightarrow{} Night} &  & \multicolumn{3}{c}{\cellcolor{pink}v.KITTI \textrightarrow{} Sem.KITTI} &  & \multicolumn{3}{c}{\cellcolor{gray}A2D2 \textrightarrow{} Sem.KITTI} \tabularnewline
    \cline{3-5} \cline{7-9} \cline{11-13} \cline{15-17} 
     &  & 2D & 3D & Avg &  & 2D & 3D & Avg &  & 2D & 3D & Avg &  & 2D & 3D & Avg \tabularnewline
    \hline 
     & Baseline (Source only) &  58.4 & 62.8 & 68.2  && 47.8 & 68.8 & 63.3 && 26.8 & 42.0 & 42.2 && 34.2 & 35.9 & 40.4 \tabularnewline
    \midrule
    \multirow{3}{*}{Uni-modal} & MinEnt \cite{vu2019advent} & 57.6 & 61.5 & 66.0 && 47.1 & 68.8 & 63.6 && 39.2 & 43.3 & 47.1 & &37.8 & 39.6 & 42.6 \tabularnewline
    &  Deep logCORAL~\cite{morerio2018minimalentropy} & 64.4 & 63.2 & 69.4 && 47.7 & 68.7 & 63.7 && 41.4 & 36.8 & 47.0 && 35.1 & 41.0 & 42.2 \tabularnewline
     & PL \cite{li2019bidirectional} & 62.0 & 64.8 & 70.4 & & 47.0 & 69.6 & 63.0 && 21.5 & 44.3 & 35.6 && 34.7 & 41.7 & 45.2\tabularnewline
    \midrule 
    \multirow{3}{*}{Multi-modal} & xMUDA \cite{jaritz2020xmuda} & 64.4 & 63.2 & 69.4 && 55.5 & 69.2 & 67.4 && 42.1 & 46.7 & 48.2 && 38.3 & 46.0 & 44.0  \tabularnewline
    & DsCML* \cite{dscml}  & 52.9 & 52.3 & 56.9 && 51.2 & 61.4 & 61.8 && 31.8 & 32.8 & 34.8 & & 25.4 & 32.6 & 33.5
    \tabularnewline
     & \algoname{} (Ours) & \textbf{71.7} & \textbf{66.8} & \textbf{72.4} && \textbf{70.5} & \textbf{70.2} & \textbf{72.1} && \textbf{53.4} &\textbf{50.3}  & \textbf{56.5} && \textbf{42.3} & \textbf{46.1} & \textbf{46.2} \tabularnewline
    \midrule
    \midrule
    & Oracle & 75.4 & 76.0 & 79.6 & & 61.5 & 69.8 & 69.2 & & 66.3 & 78.4 & 80.1 & & 59.3 & 71.9 & 73.6 \tabularnewline
    \midrule 
    \end{tabular}}
    \caption{\textbf{Results for UDA for 3D semantic segmentation with both uni-modal and multi-modal adaptation methods}. We report performance for each network stream in terms of mIoU. ‘Avg’ column denotes the obtained by taking the mean of the 2D and 3D predictions.
    * indicates trained by us using official code.
    }
    \label{table:results}
    \end{table*}

\section{Experiments}
\label{sec:experiments}
\subsection{Datasets}
\label{subsec:datasets}
To evaluate our method, we follow the benchmark introduced in \cite{jaritz2022cross} because it comprehends several interesting domain shift scenarios.
The datasets used in the benchmark are nuScenes \cite{nuscenes} A2D2 \cite{a2d2}, SemanticKITTI \cite{semanticKITTI}, and VirtualKITTI \cite{gaidon2016virtual} in which LiDAR point clouds and camera are synchronized and calibrated so that the projection between a 3D point and its corresponding 2D image pixel can always be computed.
It is important to note that only 3D points visible from the camera are used for both training and testing.
NuScenes consists of 1000 driving scenes in total, each of 20 seconds, with 40k annotated point-wise frames taken at 2Hz, and it is deployed to implement two adaptation scenarios: day-to-night and country-to-country. The former exhibits severe light changes between the source and the target domain, while the latter covers changes in the scene layout.
In both settings adaptation is performed on six classes: \textit{vehicle, driveable\_surface, sidewalk, terrain, manmade, vegetation}.
The third challenging benchmark foresees adaptation from synthetic to real data, and it is implemented by adapting from VirtualKITTI to SemanticKITTI.
Since VirtualKITTI only provides depth maps, we use the same simulated LiDAR scans from our competitor \cite{jaritz2022cross} for a fair comparison. Note also that to accommodate for the different classes in the two datasets, a class mapping is required and we use the same defined in \cite{jaritz2022cross}.
The last adaptation scenario involves A2D2 and SemanticKITTI.
The A2D2 dataset is composed of 20 drives, with a total of 28,637 frames.
As the LiDARs sensor is very sparse (16 layers), all three front LiDARs are used.
All frames of all sequences are used for training, except for the sequence 20180807\_145028 which is left out for testing.
The SemanticKITTI dataset features a large-angle front camera and a 64-layer LiDAR. Scenes from  {0, 1, 2, 3, 4, 5, 6, 9, 10} are used for training, scene 7 as validation, and 8 as a test set.
In this case, only the ten classes that are in common along the two datasets are used: \textit{car, truck, bike, person, road, parking, sidewalk, building, nature, other-objects}.
\begin{table*}[t]
    \centering{}
    \scalebox{0.62}{
    \begin{tabular}
    {cl>
    {\centering}p{1cm}>
    {\centering}p{1cm}>
    {\centering}p{1cm}>
    {\centering}p{0.1cm}>
    {\centering}p{1cm}>
    {\centering}p{1cm}>
    {\centering}p{1cm}>
    {\centering}p{0.1cm}>
    {\centering}p{1cm}>
    {\centering}p{1cm}>
    {\centering}p{1cm}>
    {\centering}p{0.1cm}>
    {\centering}p{1cm}>
    {\centering}p{1cm}>
    {\centering}p{1cm}>
    {\centering}p{0.1cm}>
    {\centering}p{1cm}>
    {\centering}p{1cm}>
    {\centering}p{1cm}>
    {\centering}p{0.1cm}>
    {\centering}p{1cm}>
    {\centering}p{1cm}>
    {\centering}p{1cm}
    }
    \hline 

    & \multirow{2}{*}{Method} & \multicolumn{3}{c}{\cellcolor{blue!25}USA \textrightarrow{} Singapore} &  & \multicolumn{3}{c}{\cellcolor{YellowOrange}Day \textrightarrow{} Night} &  & \multicolumn{3}{c}{\cellcolor{pink}v.KITTI \textrightarrow{} Sem.KITTI} &  & \multicolumn{3}{c}{\cellcolor{gray}A2D2 \textrightarrow{} Sem.KITTI} \tabularnewline
    \cline{3-5} \cline{7-9} \cline{11-13} \cline{15-17} 
     &  & 2D & 3D & Avg &  & 2D & 3D & Avg &  & 2D & 3D & Avg &  & 2D & 3D & Avg \tabularnewline
    \hline 
     & xMUDA* \cite{jaritz2020xmuda}         & 58.7 & \textbf{62.3} & 68.6 && 43.0 & \textbf{68.9} & 59.6 && 25.7 & 37.4 & 39.0 && 34.9 & \textbf{36.7} & 41.6
    \tabularnewline
    & \algoname{} (Ours)            & \textbf{69.7} & \textbf{62.3} & \textbf{70.9} && \textbf{65.3} & 63.2 & \textbf{68.3} && \textbf{37.7} & \textbf{40.2} & \textbf{44.2} && \textbf{39.6} & 35.9 & \textbf{43.6}
     \tabularnewline
    \midrule 
    \end{tabular}}
    \caption{\textbf{Results for 3D for semantic segmentation in the Domain Generalization setting.} We report performance for each network stream in terms of mIoU. ‘Avg’ column denotes the obtained by taking the mean of the 2D and 3D predictions. * indicates trained by us using official code.}
    \label{tab:results_DG}
    \end{table*}

\begin{figure*}[t]
    \centering
    \scalebox{0.7}{
    \includegraphics[width=\linewidth]{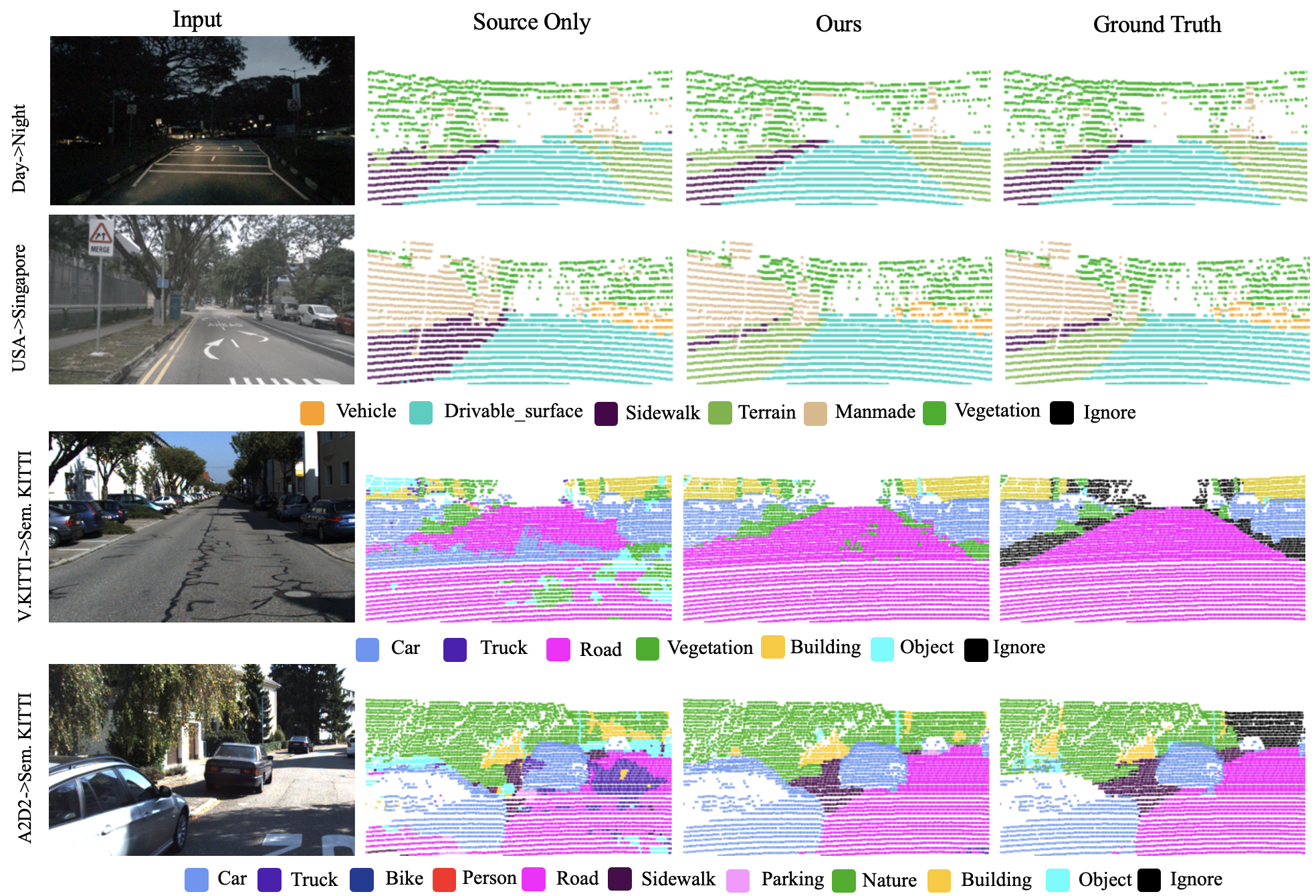}}
    \caption{\textbf{Qualitative examples of the proposed framework in the UDA scenario.} From left to right: RGB images, point cloud segmentations projected into 2D for visualization purpose of the baseline source only model, our method, and the ground truth respectively. From top to bottom: the four different adaptation scenarios. Comparisons are provided for the target domain.}
    \label{fig:qualitatives}
\end{figure*}

\subsection{Implementation details}
We use the same data augmentation pipeline as our competitors, which is composed of random horizontal flipping and color jittering for 2D images, while vertical axis flipping, random scaling, and random 3D rotations are used for the 3D scans. It is important to note that augmentations are done independently for each branch.
We implement our framework in PyTorch using two NVIDIA 3090 GPU with 24GB of RAM.
We train with a batch size of 16, alternating batches of source and target domain for the UDA case and source only in DG. The smaller dataset is repeated to match the length of the other.
We rely on the AdamW optimizer \cite{loshchilov2018decoupled} and the One Cycle Policy as a learning rate scheduler \cite{onecycle}.
We train for 50, 35, 15, and 30 epochs for USA \textrightarrow{} Singapore, Day \textrightarrow{} Night, v. KITTI \textrightarrow{} Sem. KITTI, and A2D2 \textrightarrow{} Sem. KITTI respectively.
As regards the hyper-parameters, we follow \cite{jaritz2022cross} and set
$\lambda_{s}=0.8, \lambda_{t}=0.1, \lambda_{xs}=0.1, \lambda_{xt}=0.01$ in all settings without performing any fine-tuning on these values.

\begin{table*}[t]
    \centering{}
    \scalebox{0.58}{
    \begin{tabular}
    {cl>
    {\centering}p{1cm}>
    {\centering}p{1cm}>
    {\centering}p{1cm}>
    {\centering}p{1cm}>
    {\centering}p{0.1cm}>
    {\centering}p{1cm}>
    {\centering}p{1cm}>
    {\centering}p{1cm}>
    {\centering}p{0.1cm}>
    {\centering}p{1cm}>
    {\centering}p{1cm}>
    {\centering}p{1cm}>
    {\centering}p{0.1cm}>
    {\centering}p{1cm}>
    {\centering}p{1cm}>
    {\centering}p{1cm}>
    {\centering}p{0.1cm}>
    {\centering}p{1cm}>
    {\centering}p{1cm}>
    {\centering}p{1cm}>
    {\centering}p{0.1cm}>
    {\centering}p{1cm}>
    {\centering}p{1cm}>
    {\centering}p{1cm}
    }
    \hline 

     \multirow{2}{*}{Method} &  \multirow{2}{*}{Depth} & \multirow{2}{*}{RGB} & \multicolumn{3}{c}{\cellcolor{blue!25}USA \textrightarrow{} Singapore} &  & \multicolumn{3}{c}{\cellcolor{YellowOrange}Day \textrightarrow{} Night} &  & \multicolumn{3}{c}{\cellcolor{pink}v.KITTI \textrightarrow{} Sem.KITTI} &  & \multicolumn{3}{c}{\cellcolor{gray}A2D2 \textrightarrow{} Sem.KITTI} \tabularnewline
    \cline{4-6} \cline{8-10} \cline{12-14} \cline{16-18} 
      & & & 2D & 3D & Avg &  & 2D & 3D & Avg &  & 2D & 3D & Avg &  & 2D & 3D & Avg \tabularnewline
    \hline 
     xMUDA \cite{jaritz2022cross} & & & 64.4 & 63.2 & 69.4 && 55.5 & 69.2 & 67.4 && 42.1 & 46.7 & 48.2 && 38.3 & 46.0 & 44.0  \tabularnewline
     \algoname{} (Ours)     & \checkmark &      & 69.5 & 64.0 & 69.6  && \textbf{71.3} & 69.9 & \textbf{72.8} && 52.6 & 40.3 & 53.7 && 41.7  & 44.8 & 45.9 \tabularnewline
     \algoname{} (Ours)    & \checkmark & \checkmark  & \textbf{71.7} & \textbf{66.8} & \textbf{72.4} && 70.5 & \textbf{70.2} & 72.1 && \textbf{53.4} & \textbf{50.3}  & \textbf{56.5} && \textbf{42.3} & \textbf{46.1} & \textbf{46.2}   \tabularnewline
     
    \midrule 
    \end{tabular}}
    \caption{\textbf{Modality-wise ablation of the proposed framework in the UDA scenario}. \textit{Depth} indicates the usage of the additional sparse depth encoder, while \textit{RGB} denotes the introduction of the RGB information in the 3D network.}
    \label{table:ablation_results}
    \end{table*}
    
\begin{table*}[t]
    \centering{}
    \scalebox{0.62}{
    \begin{tabular}
    {cl>
    {\centering}p{1cm}>
    {\centering}p{1cm}>
    {\centering}p{0.1cm}>
    {\centering}p{1cm}>
    {\centering}p{1cm}>
    {\centering}p{1cm}>
    {\centering}p{0.1cm}>
    {\centering}p{1cm}>
    {\centering}p{1cm}>
    {\centering}p{1cm}>
    {\centering}p{0.1cm}>
    {\centering}p{1cm}>
    {\centering}p{1cm}>
    {\centering}p{1cm}>
    {\centering}p{0.1cm}>
    {\centering}p{1cm}>
    {\centering}p{1cm}>
    {\centering}p{1cm}>
    {\centering}p{0.1cm}>
    {\centering}p{1cm}>
    {\centering}p{1cm}>
    {\centering}p{1cm}
    }
    \hline 

     \multirow{2}{*}{Method} & \multicolumn{3}{c}{\cellcolor{blue!25}USA \textrightarrow{} Singapore} &  & \multicolumn{3}{c}{\cellcolor{YellowOrange}Day \textrightarrow{} Night} &  & \multicolumn{3}{c}{\cellcolor{pink}v.KITTI \textrightarrow{} Sem.KITTI} &  & \multicolumn{3}{c}{\cellcolor{gray}A2D2 \textrightarrow{} Sem.KITTI} \tabularnewline
    \cline{2-4} \cline{6-8} \cline{10-12} \cline{14-16} 
      & 2D & 3D & Avg &  & 2D & 3D & Avg &  & 2D & 3D & Avg &  & 2D & 3D & Avg \tabularnewline
    \hline 
      \algoname{} (Ours)         & 71.7 & 66.8 & 72.4 && 70.5 & \textbf{70.2} & 72.1 && 53.4 & 50.3 & 56.5 && 42.3 & 46.1 & 46.2   \tabularnewline
    \midrule 
    \midrule 
      xMUDA \cite{jaritz2022cross} + PL & 67.0 & 65.4 & 71.2 && 57.6 & 69.6 & 64.4 && 45.8 & 51.4 & 52.0 && 41.2 & \textbf{49.8} & 47.5 \tabularnewline
      \algoname{} (Ours)  + PL & \textbf{74.3} & \textbf{68.3} & \textbf{74.9} && \textbf{71.3} & 69.6 & \textbf{72.2} && \textbf{55.4} & \textbf{55.0} & 59.7 && \textbf{46.4} & 48.7 & \textbf{50.7}  \tabularnewline
     \algoname{} (Ours)  + Fusion & x & x & 74.0 && x & x & 71.0 && x & x &\textbf{60.4} && x& x & 48.8   \tabularnewline
    
        \midrule 
    \end{tabular}}
    \caption{\textbf{Self-training Analysis.} Results with different self-training strategies in the UDA scenario.}
    \label{tab:results_st}
    \end{table*}   

\subsection{UDA results}
Following previous works in the field \cite{jaritz2022cross, dscml}, we evaluate the performance of a model on the target test set using the standard Intersection over Union (IoU) and select the best checkpoint according to a small validation set on the target domain.
In \cref{table:results}, we report our results on the four challenging UDA benchmarks explained in \cref{subsec:datasets}. For each experiment, we report two reference methods: a model trained only on the source domain, named \textit{Baseline (Source Only)}; a model trained only on the target data using annotations, representing the upper bound than can be obtained with real ground-truth, namely \textit{Oracle}. We note that these two models employ the two independent stream architecture of \cite{jaritz2022cross}.
In the columns \textit{Avg}, we report the results obtained by the mean of the 2D and 3D outputs after softmax which is the final output of our multi-modal framework.
For the sake of completeness, we also report the results of each individual branch (2D and 3D only). 
We compare our method with both Uni-modal and Multi-Modal approaches. 
In particular, we mainly focus on a comparison with xMUDA\cite{jaritz2022cross} and DsCML\cite{dscml}, as they are the current s.o.t.a. methods for UDA in our multi-modal setting.
In particular, for the latter, we use the official code provided by the authors \footnote{\url{https://github.com/leolyj/DsCML}} to retrain the model on the new more exhaustive benchmark defined by \cite{jaritz2022cross}.
Overall, we note how our contributions largely improve results over competitors across all settings and modalities.
In USA \textrightarrow{} Singapore, we observe a large boost in both branches, and on average we report a +3\% (third row of the Multi-modal section). The large improvement (+7.3\%) for the 2D model, suggests that the depth cues injected into a common 2D decoder can be quite useful even if the light conditions are similar across domains. 
In Day \textrightarrow{} Night, we observe a remarkable +15\% for the 2D branch, which in turn rises the average score to +4.7\% when compared with the previous best model. We attribute this boost in performance to the depth encoder, which is able to provide useful hints when the RBG encoder has to deal with large changes in light conditions.
Remarkably, our network surpasses even the performance of the two independent streams \textit{Oracle}.
Indeed, as discussed in \cref{subsec:depth_encoder}, the sparse depth is able to give useful details for the task of semantic segmentation. Moreover, thanks to the fact that the cross-modal loss \cref{subsec:crossmodal} is optimized for both domains, the network lean to use both encoders to make the final predictions, leading to more robust performance when the encoder receives a less informative RGB signal.
In the challenging synthetic-to-real case (v. KITTI \textrightarrow{} Sem. KITTI), we also notice consistent improvements in both branches. We highlight that even though RGB colors are here likely the main source of the domain gap, they are still useful to obtain a stronger 3D model (+3.6\%). 
In the A2D2 \textrightarrow{} Sem. KITTI setting, where the sensors setup is different, we still benefit from the depth hints provided to the 2D network, and on average, our method surpasses by 2.2\% xMUDA.
In general, we highlight that though we employed both modalities in the 2D and 3D branches, the Avg performances are better than those of each individual branch, supporting our core intuition.
In \cref{fig:qualitatives}, we report some qualitative results obtained with our framework.

\subsection{Domain Generalization results}
In this section, we test our contributions in the Domain Generalization setting, in which the target data cannot be used at training time. 
For this study we consider XMUDA \cite{jaritz2022cross} as our baseline two-branch 2D-3D method, and we show that our simple contribution can boost generalization performances. Results are reported in \cref{tab:results_DG}. To implement this experiment we keep the same hyper-parameters as used in the UDA scenario. We retrain \cite{jaritz2020xmuda} using the official code, but without the target data.
Also in this setting, we observe overall  large improvements. 
We believe that this can be  ascribed especially to the introduction of the depth encoder, which helps to achieve a better generalization.
Evidence of this is well observable in the Day \textrightarrow{} Night, where the 2D performance increases from 43\% to 65.3\% in terms of mIoU, but also for USA \textrightarrow{} Singapore and  (v. KITTI \textrightarrow{} Sem. KITTI), where we achieve +11\% and +12\ respectively.
In the Day \textrightarrow{} Night scenario, the 3D branch experience a drop in performance. We think that it is related to the large domain shift of RGB images. Differently from the adaptation scenario in which we can train directly on the unlabeled target data to counteract this problem, in the generalization scenario, it influences badly the 3D performance. However, we note that our final Avg prediction still outperforms xMUDA.

\subsection{Ablation Studies}
\textbf{Modality-wise analysis.}
In \cref{table:ablation_results}, we ablate our contributions starting from the model proposed by \cite{jaritz2020xmuda} in the UDA scenario.
We start by activating our depth-based network, introduced in \cref{subsec:rgb_encoder}. The performance boost given by our proposal is remarkable across all settings. In cases such as Day \textrightarrow{} Night, where the RGB gap is larger, the depth cues injected with skip connections to the semantic decoder greatly enhance performances in the target domain (+15.8\% for 2D and +5.4\% in "Avg).
We note also a consistent improvement for the remaining settings, in particular, we highlight a +10.5\% for the 2D scores on the challenging synthetic-to-real adaptation benchmark (v. KITTI \textrightarrow{} Sem. KITTI).
Furthermore, when feeding RGB colors to the 3D network (last row of \cref{table:ablation_results}), we observe improved performances in almost all settings. The largest improvement is oberved in the synthetic-to-real setting, where we achieve a +10\% in terms of mIou for the 3D, which in turn increased the average score from 53.7\% to 56.5\%. 
Better performance is also achieved for both the 3D network and the average score for A2D2 \textrightarrow{} Sem. KITTI.

\textbf{Self-Training.}
In this section, we compare different self-training strategies and report results in \cref{tab:results_st}. As explained in \cref{subsec:crossmodal}, for the self-training protocol we first need a model trained on the source domain to produce the pseudo-labels for the target domain in the second round. We report in the first row of \cref{tab:results_st} the performance of this starting model to better appreciate the effectiveness of self-training. First, we note how thanks to our contributions, for USA \textrightarrow{} Singapore, Day \textrightarrow{} Night, and v. KITTI \textrightarrow{} Sem. KITTI we already surpass xMUDA\cite{jaritz2020xmuda} on the \textit{Avg} column even without the usage of pseudo-labels. 
When pseudo-labels from the 2D and the 3D branches are used to supervise the 2D and the 3D network respectively, we establish new state-of-the-art performances for all four settings in the average predictions (third row).
Furthermore, in the fourth row of \cref{tab:results_st}, we deploy the strategy proposed in \cite{jaritz2020xmuda}, where point-wise features from the two networks are concatenated and used to train a unique classifier). 
In this case, we observe mixed results, indicating that this self-training strategy is not necessarily better across all settings when compared to the standard self-training protocol.

\section{Conclusions}
In this paper, we shed light on the complementarity of recent and emerging 3D-2D architectures for 3D semantic segmentation. 
We provide an intuitive explanation based on the notion of effective receptive field  of why processing data with these two networks grants orthogonal predictions that can be  effectively fused together. Based on this, we propose to feed both modalities to both branches. Despite the simplicity of our approach, we establish new state-of-the-art results in four common UDA scenarios and demonstrate superior generalization performance over the baseline 2D-3D architecture.
A limitation of our work is that our method is purely multi-modal, and it requires both modalities and a valid calibration across sensors at test time.
An interesting future direction is to investigate how our approach may generalize to other multi-modal  2D-3D architectures for semantic segmentation. 

{\small
\bibliographystyle{ieee_fullname}
\bibliography{egbib}
}

\end{document}